\documentclass{edm_template}

\usepackage{optidef}
\usepackage{graphicx}
\usepackage[font=small,skip=5pt]{caption}

\usepackage{subfig}

\usepackage{booktabs}
\usepackage{tabularx}
\usepackage[flushleft]{threeparttable}
\usepackage{subfloat}
\usepackage{multicol}
\usepackage{wrapfig}
\usepackage{xcolor}
\usepackage{csquotes}
\begin{document}

\title{A Self-Attentive model for Knowledge Tracing}
%
%
%
%
%

\numberofauthors{2}
\author{
\alignauthor
Shalini Pandey\\
\affaddr{University of Minnesota}\\
\affaddr{Twin Cities, MN 55455, USA}\\
\email{pande103@umn.edu}\\
\alignauthor
George Karypis\\
\affaddr{University of Minnesota}\\
\affaddr{Twin Cities, MN 55455, USA}\\
\email{karypis@umn.edu}\\
}




\maketitle

\begin{abstract}
Knowledge tracing is the task of modeling each student's mastery of knowledge concepts (KCs) as (s)he engages with a sequence of learning activities. Each student's knowledge is modeled by estimating the performance of the student on the learning activities.
  It is an important research area for providing a personalized learning platform to students.
   In recent years, methods based on Recurrent Neural Networks (RNN)  such as Deep Knowledge Tracing (DKT) and Dynamic Key-Value Memory Network (DKVMN) outperformed all the traditional methods because of their ability to capture complex representation of human learning. However, these methods face the issue of not generalizing well while dealing with sparse data which is the case with real-world data as students interact with few KCs. In order to address this issue, we develop an approach that identifies the KCs from the student's past activities that are \textit{relevant} to the given KC and predicts his/her mastery based on the relatively few KCs that it picked. Since predictions are made based on relatively few past activities, it handles the data sparsity problem better than the methods based on RNN. For identifying the relevance between the KCs, we propose a self-attention based approach, Self Attentive Knowledge Tracing (SAKT).  
  Extensive experimentation on a variety of real-world dataset shows that our model outperforms the state-of-the-art models for knowledge tracing, improving AUC by $4.43\%$ on average. 
\end{abstract}


\section{Introduction}
The availability of massive dataset of students' learning trajectories about their \textit{knowledge concepts} (KCs), where a KC can be an exercise, a skill or a concept, has attracted data miners to develop tools for predicting students' performance and giving proper feedback~\cite{self1990theoretical}. For developing such personalized learning platforms, knowledge tracing (KT) is considered to be an important task and is defined as the task of tracing a student's \textit{knowledge state}, which represents his/her mastery level of KCs, based on his/her past learning activities. The KT task can be formalized as a supervised sequence learning task - given student's past exercise interactions \( \mathbf{X} = (\mathbf{x}_1, \mathbf{x}_2, \ldots, \mathbf{x}_t) \), predict some aspect of his/her next interaction $\mathbf{x}_{t+1}$. On the question-answering platform, the interactions are represented as
$\mathbf{x}_t = (e_t, r_t)$, where \( e_t \) is the exercise that the student attempts at timestamp $t$ and $r_t$ is the correctness of the student's answer. KT aims to predict whether the student will be able to answer the next exercise correctly, i.e., predict \( p(r_{t+1}=1| e_{t+1}, \mathbf{X}) \).\par
Recently deep learning models such as Deep Knowledge Tracing (DKT)~\cite{piech2015deep} and its variant~\cite{yeung2018addressing} used Recurrent Neural Network (RNN) to model a student's knowledge state in one summarized hidden vector. 
Dynamic Key-value memory network (DKVMN)~\cite{zhang2017dynamic} exploited Memory Augmented Neural Network~\cite{santoro2016one} for KT. Using two matrices, \textit{key} and \textit{value}, it learns the correlation between the exercises and the underlying KC and student's knowledge state, respectively.
The DKT model faces the issue of its parameters being non-interpretable ~\cite{khajah2016deep}. 
DKVMN is more interpretable than DKT as it explicitly maintains a KC representation matrix (\textit{key}) and a knowledge state representation matrix (\textit{value}). However, since all these deep learning models are based on RNNs, they face the issue of not generalizing while dealing with sparse data ~\cite{Kang2018SelfAttentiveSR}.  \par


\begin{figure}

\includegraphics[keepaspectratio,width=0.5\textwidth]{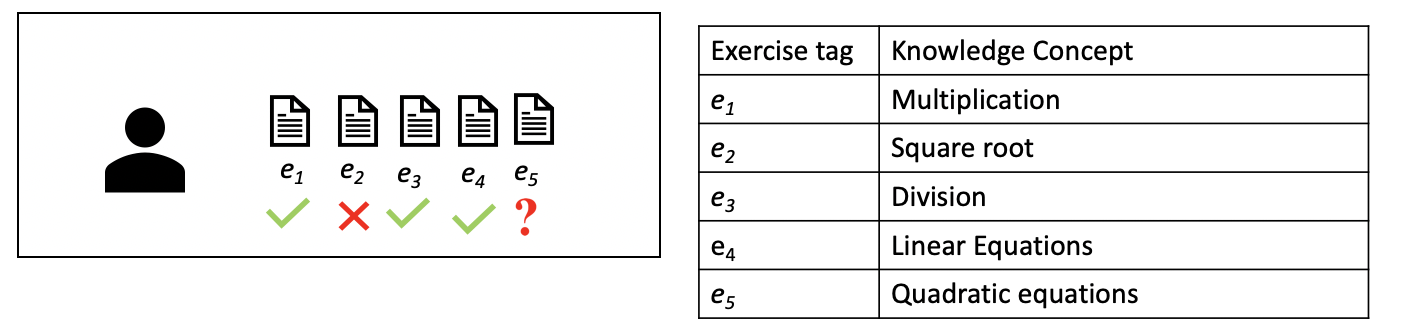}
\caption{  Left subfigure shows the sequence of exercises that the student attempts and the right subfigure shows the knowledge concepts to which each of the exercises belong.}
 \label{first}
\end{figure}
In this paper, we propose to use a purely attention mechanism based method, \textit{transformer}~\cite{vaswani2017attention}.  
In the KT task, the skills that a student builds while going through the sequence of learning activities, are related to each other and the performance on a particular exercise is dependent on his performance on the past exercises related to that exercise. For example, in figure~\ref{first}, for a student to solve an exercise on \enquote{Quadratic equation} (exercise 5) which belongs to the knowledge concept \enquote{Equations}, he needs to know how to find \enquote{square roots} (exercise 3) and \enquote{linear equations} (exercise 4). SAKT, proposed in this paper first identifies \textit{relevant} KCs from the past interactions and then predicts student's performance based on his/her performance on those KCs. For predicting student's performance on an exercise, we used exercises as KCs. As we show later,  SAKT assigns weights to the previously answered exercises, while predicting the performance of the student on a particular exercise.   
The proposed SAKT method significantly outperforms the state-of-the-art KT methods gaining a performance improvement of $4.43\%$ on the AUC, on an average across all datasets. Furthermore, the main component (self-attention) of SAKT is suitable for parallelism; thus, making our model order of magnitude faster than RNN based models.   \par

   \begin{figure}[!ht]
     \subfloat[Network of SAKT. At each timestamp the attention weights are estimated for each of the previous element only. Keys, Values and Queries are extracted from the embedding layer shown below. When $j$th element is query and $i$th element is key, attention weight is $a_{i,j}$. \label{subfig-1}]{%
       \includegraphics[width=0.45\textwidth]{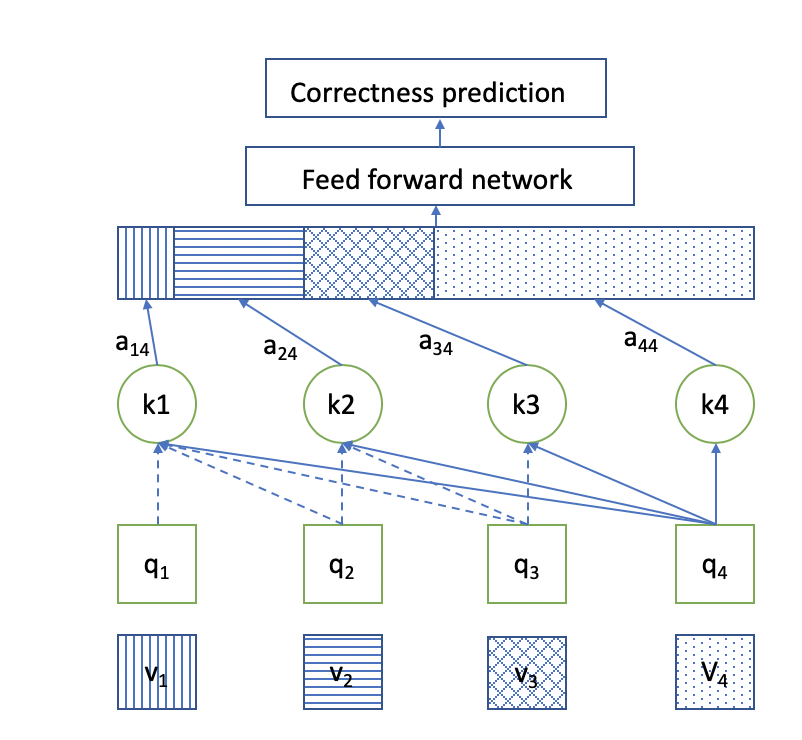}
     }
    
     \subfloat[ Embedding layer embeds the current exercise that the student is attempting and his past interactions. At every time stamp $t+1$, the current question $e_{t+1}$ is embedded in the query space using Exercise embedding and elements of past interactions $\textbf{x}_t$ is embedded in the key and value space using the Interaction embedding.  
     \label{subfig-2}]{%
       \includegraphics[width=0.45\textwidth]{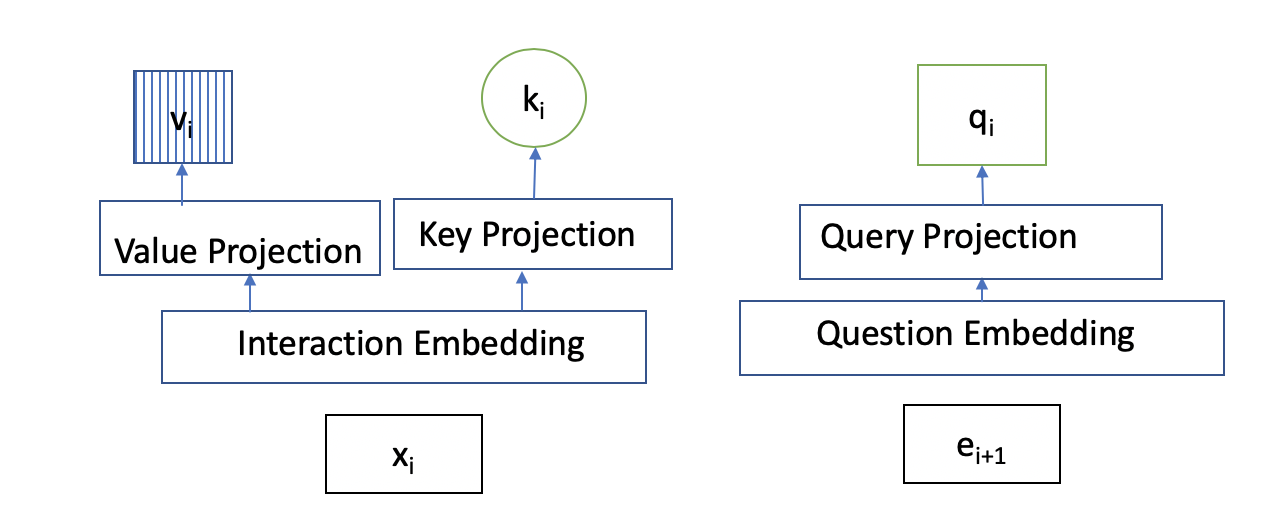}
     }
    
     \caption{Diagram showing the architecture of SAKT.}
     \label{fig:dummy}
   \end{figure}
\def\hat{\mathaccent "705E\relax}
\begin{table}[]
\caption{Notations}
\label{notations}
\begin{tabular}{ll}
\toprule
Notations & Description\\
\hline
$N$                                                  & total number of students                                                     \\
$E$                                                  & total number of exercises                                                    \\
$\textbf{X}$                        & Interaction sequence of a student: $(x_1, x_2, \ldots, x_t)$ \\
$x_i $                    & $i$th exercise-answer pair of a student                                      \\
$n$                                                  & maximum length  of sequence                                                  \\
$d$                                                 & latent vector dimensionality                                                 \\
$\textbf{e} $                       & Sequence of exercises solved by the student                                  \\
$\textbf{M} $                       & Interaction embedding matrix                                        \\
$\textbf{P} $                      & Positional embedding matrix                                                  \\
$\textbf{E}   $                     & Exercise lookup matrix                                                     \\
$\hat{\textbf{M}}$ & Past interactions embedding                                                  \\
$\hat{\textbf{E}}$ & Exercise embedding                                \\
\bottomrule
\end{tabular}
\end{table}

\section{Proposed Method}

Our model predicts whether a student will be able to answer the next exercise $e_{t+1}$ based on his previous interaction sequence $\textbf{X}={\textbf{x}_1, \textbf{x}_2, \ldots, \textbf{x}_t}$.  As shown in figure 2, we can transform the problem into a sequential modeling problem. It is convenient to consider the model with inputs ${\textbf{x}_1, \textbf{x}_2, \ldots, \textbf{x}_{t-1}}$ and the exercise sequence with one position ahead, ${e_2, e_3, \ldots, e_t}$ and the output being the correctness of the response to exercises ${r_2, r_3, \ldots, r_t}$. 
 The interaction tuple \( \textbf{x}_t = (e_t,r_t) \) is presented to the model as a number $y_t = e_t+ r_t\times E$, where $E$ is the total number of exercises. Thus, the total values that an element in the interaction sequence can take is $2E$, while elements in the exercise sequence can take $E$ possible values.   \par
 We now describe the different layers of our architecture.\par
 \textbf{Embedding layer:}
 We transform the obtained input sequence $\textbf{y} = (y_1, y_2, \ldots, y_t)$ into $s = (s_1, s_2, \ldots , s_n)$, where $n$ is the maximum length that the model can handle. Since the model can work with inputs of fixed length sequence,  if the sequence length, $t$ is less than $n$,  we repetitively add a \textit{padding} of question-answer pair to the left of the sequence. However, if $t$ is greater than $n$, we partition the sequence into subsequences of length $n$. Specifically, when $t$ is greater than $n$, $y_t$ is partitioned into $t/n$ subsequences each of length $n$. All these subsequences serve as input to the model. \\
We train an \textit{Interaction embedding matrix}, ${\textbf{M}}\in \mathbb{R}^{2E\times d}$, where $d$ is the latent dimension. This matrix is used to obtain an embedding, $\textbf{M}_{s_i}$ for each element, $s_i$ in the sequence. 
Similarly, we train exercise embedding matrix, ${\textbf{E}} \in \mathbb{R}^{E\times d}$ such that each exercise in the set $e_i$ is embedded in the ${e_i}$th row.\\
\textit{Position Encoding:}
Position Encoding is the layer in the self-attention neural network which is used for encoding the position so that like convolution network and recurrent neural network, we can encode the order of the sequence. This layer is particularly important in knowledge tracing problem because a student's knowledge state evolves gradually and steadily with time. 
The knowledge state at a particular time instance should not show wavy transitions~\cite{yeung2018addressing}. In order to incorporate this we use a parameter, position embedding, $\textbf{P} \in \mathbb{R}^{n\times d}$ which is learned while training. The $i$th row of position embedding matrix, $\textbf{P}_i$ is then added to the interaction embedding vector of the $i$th element of the interaction sequence. \par
The output from the embedding layer is embedded interaction input matrix, $\hat\textbf{M}$ and embedded exercise matrix, $\hat\textbf{E}$: 
\begin{equation}
    \hat\textbf{M} = \begin{bmatrix}
       \textbf{M}_{s_1}+\textbf{P}_1     \\[0.5em]  
       \textbf{M}_{s_2}+\textbf{P}_2  \\[0.5em]  
       \ldots \\[0.5em]  
       \textbf{M}_{s_n}+\textbf{P}_n 
       \end{bmatrix}, \  \hat\textbf{E} = \begin{bmatrix}
       \textbf{E}_{s_1}   \\[0.5em]     
       \textbf{E}_{s_2}   \\[0.5em]  
       \ldots \\[0.5em]  
       \textbf{E}_{s_n}
       \end{bmatrix}.
\end{equation}

\textbf{ Self-attention layer:}
In our model, we use the scaled dot-product attention mechanism ~\cite{vaswani2017attention}. This layer finds the relative weight corresponding to each of the previously solved exercise for predicting the correctness of the current exercise. 

We obtain query and key-value pairs using the following equations:
\begin{equation}
    \textbf{Q} =  \hat{\textbf{E}} \textbf{W}^{Q}, \textbf{K} =  \hat{\textbf{M}} \textbf{W}^{K}, \textbf{V} =  \hat{\textbf{M}} \textbf{W}^{V},
\end{equation}
where $\textbf{W}^{Q}$, $\textbf{W}^K$, $\textbf{W}^V  \in \mathbb{R}^{d\times d}$ are the query, key and value projection matrices, respectively, which linearly project the respective vectors to different space ~\cite{vaswani2017attention}. 
The relevance of each of the previous interactions with the current exercise is determined using the attention weights. For finding the attention weights we use the scaled dot product~\cite{vaswani2017attention}, defined as: 
\begin{equation}
  \text{Attention}(\textbf{Q}, \textbf{K}, \textbf{V}) = \text{softmax}\bigg(\frac{\textbf{Q}\textbf{K}^T}{\sqrt{d}}\bigg) \textbf{V}.
\end{equation}
\\
\textit{Mutiple heads:}
In order to jointly attend to information from different representative subspaces, we linearly project the queries, keys and values $h$ times using different projection matrices.
\begin{equation}
    \text{Multihead}(\hat\textbf{M}, \hat\textbf{E}) = \text{Concat}(\text{head}_1, \ldots, \text{head}_h) \textbf{W}^O,
\end{equation}
\noindent where $\text{head}_i=\text{Attention}(\hat\textbf{E} \textbf{W}_i^Q, \hat\textbf{M} \textbf{W}_i^K, \hat\textbf{M}\textbf{W}_i^V)$ and $\textbf{W}^O\in \mathbb{R}^{hd \times d}$.    \\
\textit{Causality:}\\
In our model, we should consider only first $t$ interactions when predicting the result of the $(t+1)$st exercise. Therefore, for a query $\textbf{Q}_i$, the keys $\textbf{K}_j$ such that $j>i$ should not be considered. We use, causality layer to mask the weights learned from a future interaction key,

\textbf{Feed Forward layer:}\\
The self-attention layer described above results in weighted sum of values, $\textbf{V}_i$ of the previous interactions. However the rows of the matrix obtained from the multihead layer, $\textbf{S} = \text{Multihead}(\hat{\textbf{M}}, \hat{\textbf{E}}) $ is still a linear combination of the values, $\textbf{V}_i$ of the previous interactions. To incorporate non-linearity in the model and consider the interactions between different latent dimensions, we use a feed forward network. 
\begin{equation}
    \textbf{F}= \text{FFN}(\textbf{S}) = \text{ReLU}(\textbf{S}\textbf{W}^{(1)} + \textbf{b}^{(1)}) \textbf{W}^{(2)}+\textbf{b}^{(2)},
\end{equation}
\noindent where $\textbf{W}^{(1)}\in \mathbb{R}^{d\times d}$, $\textbf{W}^{(2)} \in \mathbb{R}^{d\times d}$, $\textbf{b}^{(1)} \in \mathbb{R}^d$, $\textbf{b}^{(2)} \in \mathbb{R}^d $ are parameters learned during training. \\
\\
\textbf{Residual Connections:}
The residual connection ~\cite{he2016deep} 
  are used to propagate the lower layer features to the higher layers. Hence, if low layer features are important for prediction, the residual connection will help in  propagating them to the final layers where the predictions are performed.  In the context of KT, students attempt exercises belonging to a specific concept to strengthen that concept. Hence, residual connection can help propagating  the embeddings of the recently solved exercises to the final layer making it easier for model to leverage the low layer information.  A residual connection is applied after both self-attention and feed forward layer. \\
\\
\textbf{Layer normalization:} 
In~\cite{ba2016layer}, it was shown that normalizing inputs across features can help in stabilizing and accelerating neural networks. We used layer normalization in our architecture for the same purpose.
Layer normalization is also applied at both the self-attention and feed forward layer. \par
\textbf{Prediction layer:}\\
Finally, each row of the matrix $\textbf{F}_i $ obtained above is passed through the fully connected network with Sigmoid activation to predict the performance of the student. 
\begin{equation}
    p_i = \text{Sigmoid}(\textbf{F}_i\textbf{w}+\textbf{b}),
\end{equation}
where $p_i$ is a scalar and represents the probability of student providing correct response to exercise $e_i$, $\textbf{F}_i$ is the $i$th row of $\textbf{F}$ and Sigmoid(z) = $1/(1+e^{-z})$\\
\\
\textbf{Network Training:} 
The objective of training is to minimize the negative log likelihood of the observed sequence of student responses under the model. The parameters are learned by minimizing the cross entropy loss between $p_t$ and $r_t$.
\begin{equation}
    \mathcal{L} = -\Sigma_t (r_t \log(p_t)+ (1-r_t)\log(1-p_t))
\end{equation}


\section{Experimental Settings}
\subsection{Datasets}
To evaluate our model, we used four real-world datasets and one synthetic dataset.
\vspace{-4mm}
\begin{itemize}
\item \textit{Synthetic\footnote{https://github.com/chrispiech/DeepKnowledgeTracing/tree/\\master/data/synthetic}:} This dataset is obtained by simulating 4000 virtual students' answering trajectories. Each student answers the same sequence of 50 exercises, which are drawn from $5$ virtual concepts with varying difficulty level.
\vspace{-1mm}
     \item \textit{ASSISTment 2009\footnote{https://sites.google.com/site/assistmentsdata/home/assistment-2009-2010-data/skill-builder-data-2009-2010} (ASSIST2009):} This dataset is provided by ASSISTment online tutoring platform and is widely used for KT tasks.  We conducted our experiments on the updated "skill-builder" dataset. 
     The dataset is sparse as the density of this dataset is $0.06$, shown in Table 2.
     \vspace{-1mm}
\item \textit{ASSISTment 2015\footnote{https://sites.google.com/site/assistmentsdata/home/2015-assistments-skill-builder-data} (ASSIST2015):}ASSISTment 2015 contains students' responses on 100 skills. There are 19,917 students and 708,631 interactions. Although the number of records in this dataset is more than ASSISTment 2009, the average number of records per student is smaller because the number of students is larger. This dataset is the most sparse of all the available datasets, with a density of $0.05$. 
\vspace{-1mm}
\item \textit{ASSISTment Challenge (ASSISTChall):} This data is obtained from ASSISTment 2017 competition\footnote{https://sites.google.com/view/assistmentsdatamining}. It is the richest dataset in terms of the number of interactions with 942,816 interactions, 686 students and 102 skills. This dataset is the most dense dataset of all the available datasets because its density is $0.81$. 
\vspace{-1mm}
\item \textit{STATICS2011 (STATICS):} This dataset contains the interaction from an engineering statics course with 189,927 interactions, 333 students and 1223 skill tags. We adopted the processed data from~\cite{zhang2017dynamic}. It is also a dense dataset with a density of 0.31.
\end{itemize}
\vspace{-4mm}
The complete statistical information for all the datasets can be found in Table 2.
\begin{table}[]
\centering
\label{dataset}
\caption{Dataset Statistics}
\fontsize{7}{12}\selectfont
\setlength\tabcolsep{1.6pt}
\newcommand*{\thead}[1]{%
\multicolumn{1}{c}{\bfseries\begin{tabular}{@{}c@{}}#1\end{tabular}}}
  \begin{threeparttable}
\begin{tabular}{lrrrrrrr}

 \toprule
\thead{Datasets }            & \thead{\#Users} & \thead{\#Skill\\ tags } & \thead{\#Interactions} &\thead{\#Unique\\ Interactions}& \thead{Density}\\
\midrule
     
Synthetic-5          & 4000     & 50  &200K & 200K  & 1                    \\
ASSIST2009      & 4417     & 124           & 328K & 35K &0.06                              \\
ASSIST2015      & 19917    & 100           & 709K & 102K & 0.05            \\
ASSIST-Chall & 686      & 102           & 943K & 57K    &0.81          \\
STATICS             & 333      & 1223          & 190K & 129K &  0.31 \\

\bottomrule
\end{tabular}
\begin{tablenotes}[para]
The columns corresponding to \#Users, \#Skill tags and \#Interactions represent the number of students, total number of exercise tags and the number of records, respectively.
 The column Density represents the density of each dataset (i.e., $\text{Density} = \text{\#Unique Interactions}/(\text{\#Users}\times\text{\#Skill tags}$)).

\end{tablenotes}
  \end{threeparttable}
\end{table}
\subsection{Evaluation Methodology}
\textbf{Metrics:} The prediction task is considered in a binary classification setting i.e., answering an exercise correctly or not. Hence, we compare the performance using the Area Under Curve (AUC) metric. \\
\textbf{Approaches:}
We compare our model against the  state-of-the-art KT methods, DKT~\cite{piech2015deep}, DKT+ ~\cite{yeung2018addressing}, and DKVMN~\cite{zhang2017dynamic}. These methods are described in the introduction. \\
\textbf{Model Training and parameter selection:}
We trained the model with $80\%$ of the dataset and test it on the remaining. For all the methods, we tried the hidden state dimension $d = \{  50, 100, 150, 200\}$.  
 For the competing approaches, we used the same hyperparameters as reported in their respective papers. 
 For initialization of weights and optimization, we used a similar procedure as~\cite{yeung2018addressing}.  We implemented SAKT with \textit{Tensorflow} and used ADAM~\cite{kingma2014adam} optimizer with learning rate of $0.001$. We used a batch size of $256$ for the ASSISTChall dataset and $128$ for the others. For datasets with a larger number of records, e.g., ASSISTChall and ASSIST2015, we used a dropout rate of 0.2, while for the remaining datasets, we used a dropout rate of $0.2$. We set the maximum length of the sequence, $n$ as roughly proportional to the average exercise tags per student.  For ASSISTChall and STATICS dataset we use $n=500$, for the ASSIST2009  $n=100$ and $50$ , for the synthetic and ASSIST2015  datasets $n$ is set to $50$.

 \begin{table}[]
\label{performance}
\fontsize{7}{12}\selectfont
\newcommand*{\thead}[1]{%
\multicolumn{1}{c}{\bfseries\begin{tabular}{@{}c@{}}#1\end{tabular}}}
\caption{Student Performance prediction comparison.    }
\centering
\begin{threeparttable}
\begin{tabular}{lrrrrr}
\toprule
{Datasets}             & \multicolumn{5}{c}{{AUC}}                                         \\
\hline
\multicolumn{1}{l}{} & \thead{DKT}   &\thead{ DKT+ }          & \thead{DKVMN} & \thead{SAKT}           & \thead{Gain\%} \\
\cmidrule{2-6}
Synthetic          & 0.823 & 0.824 & 0.822 & \textbf{0.832}          & 0.97         \\
ASSIST2009      & 0.820 & 0.822          & 0.816 & \textbf{0.848} & 3.16          \\
ASSIST2015      & 0.736 & 0.737          & 0.727 & \textbf{0.854} & 15.87         \\
ASSISTChall & \textbf{0.734} & 0.728          & 0.689 &\textbf{0.734} & 0.00          \\
STATICS              & 0.815 & 0.835          & 0.814 & \textbf{0.853} & 2.16 \\
\hline
Average & 0.786 & 0.789          & 0.773 &\textbf{0.824} & 4.43          \\

\bottomrule
\end{tabular}
\begin{tablenotes}
\item[1] \textbf{Bold} numbers are the best performance.
\item[2] The reported results are obtained by the best hyperparameter selection for each dataset individually. 

\end{tablenotes}
\end{threeparttable}
\end{table}
 \newpage
\section{Results and Discussion}
\textbf{Student Performance Prediction:}
 Table 3 shows the performance comparison of SAKT with the current state-of-the-art methods. On the Synthetic dataset, SAKT performs better than the competing approaches, achieving an AUC of $0.832$ compared to $0.824$ by DKT+.  Even though Synthetic is the most dense dataset, SAKT outperforms RNN based methods because of the methodology used for generating Synthetic. 
 For this dataset, each individual exercise is derived from only one concept. The probability of a student answering an exercise from this dataset correctly is determined using Item Response Theory~\cite{self1990theoretical} as,
$
    p(\text{correct}| \alpha, \beta) = c + \frac{1-c}{1+\exp(\beta-\alpha)}, 
$
 where $c$ denotes the probability of guessing it correctly,  $\alpha$ and $\beta$ are randomly chosen numbers to indicate the concept ability and exercise difficulty, respectively. Thus, in this dataset, the exercises belonging to the same concept are strongly correlated. SAKT, unlike other benchmarks, directly attempts to identify exercises belonging to the same concept and hence performs better than other methods. On  ASSIST2009, SAKT performs better than competing approaches, gaining a performance improvement of $3.16\%$ over the second best performing method.  For ASSIST2015 dataset, SAKT shows an impressive improvement of $15.87\%$. We attribute this gain to the fact that attention mechanism leveraged by SAKT can learn and generalize well even when the dataset is sparse, which is the case with ASSIST2015 as its density is the least among the other datasets. For STATICS2011, our method achieves a performance improvement of $2.16\%$ compared to DKT+. For ASSISTChall, our method performs at par with DKT. This can be attributed to the fact that ASSISTChall is the most dense dataset of all the real-world datasets.

 \begin{figure}[]

     \subfloat[Heatmap depicting the attention weights between each pair of exercises. Note that, the weight assigned for pair $(i,j)$, where $j>i$ is always zero because all the sequences consists of exercises in the same order from 
     \label{heatmap1}]{%
       \includegraphics[width=0.5\textwidth]{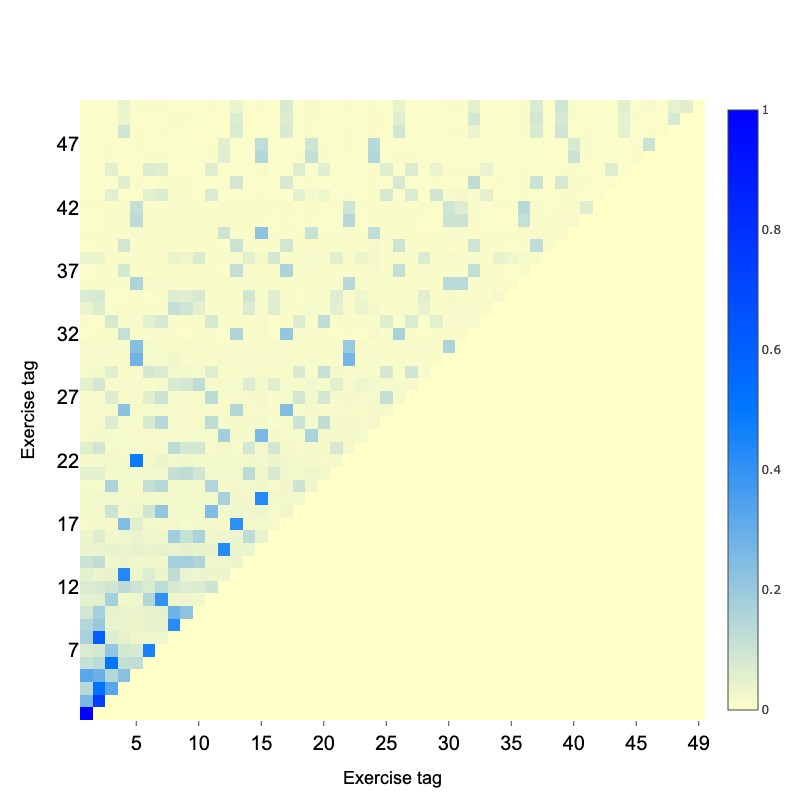}
     }
    
     \subfloat[Graph depicting the relevance between exercises. The relevance is determined by the attention weights learned between the exercises using SAKT. We observe a perfect clustering of latent concepts.  \label{heatmap2}
     ]
   {
       \includegraphics[width=0.45\textwidth]{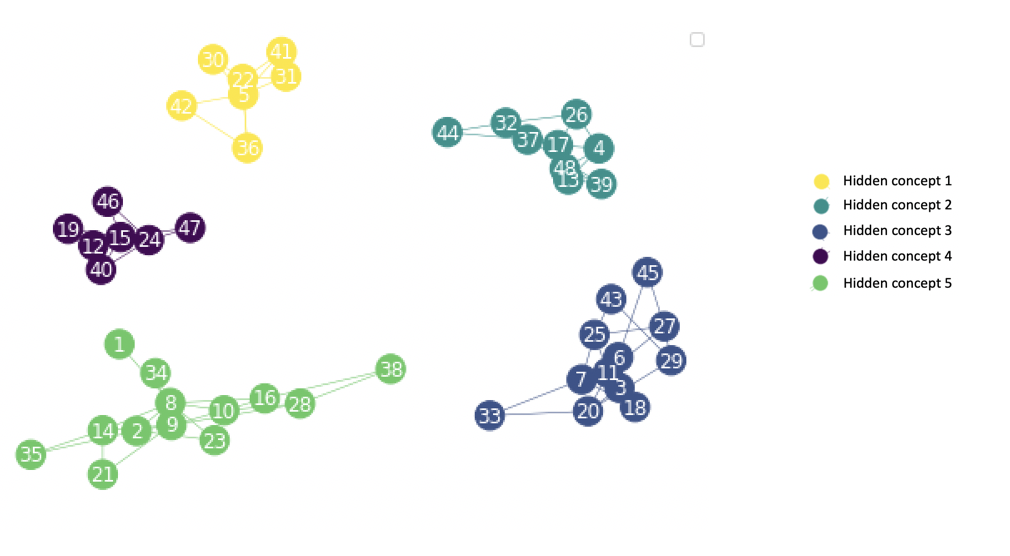}
     }
    
     \caption{Visualizing attention weight of Synthetic dataset.}
     
     \label{heatmap}
   \end{figure}
   
  
\begin{table*}[]
\caption{Example of attention weights for some sequences in ASSIST2009 dataset. }
\centering
\newcommand*{\thead}[1]{\multicolumn{1}{c}{\bfseries #1}}
  \begin{threeparttable}
\begin{tabular}{p{3cm} p{14cm}}
\toprule

\thead{Exercise tag   }                       & \thead{Past Interactions }                                          \\ \midrule

Scale Factor                          & (Probability of Two Distinct Events,1): 0.000001, (Circle Graph, 1): 0.0001, (Circle Graph,1):0.001, {\color[HTML]{FE0000}(Division Fractions, 0): 0.99}     \\

\hline

Ordering integers                     & (Intercepts,0): 0.21, {\color[HTML]{FE0000}(Ordering positive decimals,1): 0.611}, (Multiplication whole numbers,1): 0.09, (Proportion,1):0.033     \\
\hline

Rate                                  & (Interior Angles Figures, 0):0.005, (Algebraic Simplification,0) : 0.009, {\color[HTML]{FE0000}(Rate,0):0.5}, (Interior Angles Figures, 0):0.1, (Algebraic Simplification,0) : 0.12 \\

\bottomrule                                         
\end{tabular}
\begin{tablenotes}
\begin{itemize}
    \item The columns corresponding to Exercise tag refers to the query (i.e., the exercise for which we have to predict the student's performance) and Past Interactions refers to the sequence of interactions that has been observed for that student, respectively.
    \item {\color[HTML]{FE0000}red} colored elements in the right column represent the most important element among the past interaction elements.
\end{itemize}
 
\end{tablenotes}
  \end{threeparttable}
\end{table*}
  \vspace{-4mm}
 \textbf{Attention weights visualization:}
 Visualizing the attention weights between the elements of past interactions (which serve as keys) and the exercise that the student is going to solve next (which serves as query) can help in understanding which exercises in the past interactions are relevant to the query exercise. With this motivation, we compute the sum of attention weights of each exercise pair $(e1,e2)$ across all the sequences where $e1$ serves as query and interaction with exercise $e2$ serves as key. We then normalize the attention weights so that the sum of the weights for each query is one. This results in a $relevance$ matrix in which each element, $(e1,e2)$ represents the influence of $e2$ on $e1$. We perform our analysis on Synthetic because this dataset was generated with known hidden concepts and hence the ground truth regarding the relevance of different exercises are known to us. Figure~\ref{heatmap1} shows the heatmap corresponding to the $relevance$ matrix of exercises in Synthetic. For Synthetic, all the sequences consist of all exercise tags in the same sequence starting from $1$ to $50$.  \\
 In order to build the influence graph between the exercise tags, as shown in Figure~\ref{heatmap2}, we use the $relevance$ matrix. Firstly, we draw out the first exercise in the sequence that belongs to each hidden concept, and visit each row of the $relevance$ matrix, and connect the exercise corresponding to that row to the first two exercises ranked based on edge weight, which is proportional to the attention weights between the pair of exercises. We can see that the based on the attention weights,  we are able to achieve the perfect clustering of the exercise tags based on the hidden concepts from which they are derived.  An interesting observation is that two exercises which occur far apart in the sequence but belonging to the same concept can be identified by SAKT. For example, as shown Figure~\ref{heatmap2} a query on exercise 22 assigned most weight to the key with exercise 5 even when they occur far apart in the sequence. \\
  Two exercises which are relevant to each other tend to have high attention weights as the performance on one of them impacts the performance on the other. Additionally, in the real-world scenario, the exercises which occur close in the sequence tend to belong to the same concept. Thus, we expect that the attention weights biased towards the exercises that occur recently in the interaction sequence. 
  To illustrate this, we manually analyzed ASSIST2009 dataset to visualize the attention weights for some selected samples. Table 4 shows some of the exercises along with the past interactions and attention weights assigned to each interaction.  
\\
\textbf{Ablation Study:}
Table 4 shows the performance of default SAKT architecture and all the variants on all the datasets (with $d=200$). \\
\textit{No Positional Encoding (PE): } In this variant of the default architecture, we removed the positional encoding. As a result, the attention weights assigned for predicting the performance of student on a particular exercise depends only on the interaction embedding, without being affected by its position in the sequence. In case of ASSIST2009 and ASSIST2015, the dataset is sparse and hence the impact of removal of PE is not much pronounced as is the case with the dense dataset such as ASSISTChall and STATICS.\\ 
\textit{No Residual Connection (RC):}RCs shows the importance of low level features i.e., the interaction embedding while making the prediction. Since our architecture is not  very deep, 
the RC do not contribute much to the performance of the model. In fact removal of residual connection gives better performance than default for the ASSIST2015 dataset.\\
\textit{No Dropout:}
Dropout is used in neural network to regularize the model so that it can generalize better. Overfitting of the model is more effective for dataset with less number of records compared to the number of parameters of model. As a result, role of dropout is more effective for ASSIST2009 dataset and STATICS dataset.\\
\textit{Single head:}
Instead of using $5$ heads as is the case in default architecture, we tried a variant of using only one head. Multiple heads help in capturing the attention weights in different subspaces. Using single head consistently drops the performance of SAKT on all the datasets.  \\
\textit{No block:} 
When no self-attention block is used the prediction of the next exercise depends only on the last interaction. It can be seen that without attention block the performance is significantly worse than that of default architecture.  \\
\textit{2 Blocks:}
Increasing the number of blocks of self-attention increases the number of parameters of the model.  However, in our case this increase of parameters does not prove to be useful in improving the performance. The reason being an important aspect of prediction of performance of student at an exercise is dependent on his performance on the past relevant exercises. Adding another block of self-attention makes the model more complex.

\begin{figure}[!ht]
    
      \includegraphics[width=0.45\textwidth]{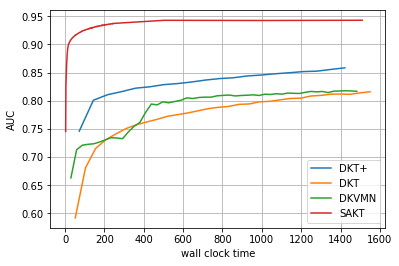}
      \vspace{-1mm}
     \caption{Training Efficiency on ASSIST2009 dataset. 
     }
     \label{fig:tr_eff}
  \end{figure}

\textbf{Training efficiency:}
Figure~\ref{fig:tr_eff} demonstrates the efficiency of various methods based on their run times on GPU during the training phase. Comparing the computational efficiency, SAKT only spends 1.4 seconds in one epoch which is 46.42 less than the time taken by DKT+ (65 seconds/epoch), 32 times less than DKT (45 seconds/epoch) and 17.33 times less than DKVMN (26 seconds/epoch).  We conducted the experiments on a single GPU of type NVIDIA Titan V. 
\begin{table}[t]
\centering
\label{dataset}
\caption{Ablation Study}
\fontsize{7}{12}\selectfont
\setlength\tabcolsep{1.9pt}
\newcommand*{\thead}[1]{%
\multicolumn{1}{c}{\bfseries\begin{tabular}{@{}c@{}}#1\end{tabular}}}

\begin{tabular}{lccccc}

 \toprule
\thead{Architecture }            & \thead{Synthetic} & \thead{ASSIST\\2009} & \thead{ASSIST\\2015} & \thead{ASSIST\\Chall} & \thead{STATICS} \\
\midrule

    Default         &      \textbf{0.832}     & \textbf{0.848}&0.854          & \textbf{0.734}  & \textbf{0.853}           \\
No PE      &   0.827    & 0.842           & 0.849 & 0.715                    & 0.832  \\        
No RC      & 0.823    & 0.847          & \textbf{0.857} & 0.709                      & 0.834                   \\
No Dropout & 0.832      & 0.845          & 0.851  & 0.711                      & 0.840               \\
Single head             & 0.823      & 0.828          & 0.845  & 0.709                      & 0.851  \\
0 block      & 0.826     & 0.837          & 0.822  & 0.634                      & 0.819  \\
2 blocks     & 0.827      & 0.840          & 0.853  & 0.724                    & 0.845  \\

\bottomrule
\end{tabular}
\end{table}

\section{Conclusion and future work}

In this work, we proposed a self-attention based knowledge tracing model, SAKT. It models a student's interaction history (without using any RNN) and predicts his performance on the next exercise by considering the relevant exercises from his past interactions. Extensive experimentation on a variety of real-world datasets shows that our model can outperform the state-of-the-art methods and is an order of magnitude faster than the RNN-based approaches.
\par

\vspace{-1mm}
\bibliographystyle{SIGCHI-Reference-Format}
\bibliography{main}

\end{document}